\documentclass{article}

\usepackage{microtype}
\usepackage{graphicx}
\usepackage{subfigure}
\usepackage{booktabs} 

\usepackage{hyperref}

\usepackage[accepted]{tom2023}

\usepackage{amsmath}
\usepackage{amssymb}
\usepackage{mathtools}
\usepackage{amsthm}
\usepackage{csquotes}
\usepackage{enumitem}

\usepackage[capitalize,noabbrev]{cleveref}

\theoremstyle{plain}

\theoremstyle{definition}

\theoremstyle{remark}

\usepackage[textsize=tiny]{todonotes}

\usepackage{amsmath,amsfonts,bm}

\def\eqref#1{equation~\ref{#1}}

\def\1{\bm{1}}

\DeclareMathAlphabet{\mathsfit}{\encodingdefault}{\sfdefault}{m}{sl}
\SetMathAlphabet{\mathsfit}{bold}{\encodingdefault}{\sfdefault}{bx}{n}

\def\gU{{\mathcal{U}}}

\DeclareMathOperator*{\argmax}{arg\,max}
\DeclareMathOperator*{\argmin}{arg\,min}

\renewcommand{\sectionautorefname}{\S\kern-0.2em}
\renewcommand{\subsectionautorefname}{\S\kern-0.2em}

\newcommand{\reallistener}{L_{\textrm{real}}}
\newcommand{\tomlistener}{L_{\textrm{ToM}}}

\newcommand{\unboundspeaker}{S_{\textrm{ups}}}
\newcommand{\boundspeaker}{S_{\textrm{bps}}}
\newcommand{\basespeaker}{S_{\textrm{base}}}

\newcommand{\utterance}{u}
\newcommand{\intention}{z}
\newcommand{\context}{c}

\icmltitlerunning{Understanding RLHF from a Bayesian Cognitive Modeling Perspective}

\begin{document}

\twocolumn[
\icmltitle{Language Models are Bounded Pragmatic Speakers: \\ Understanding RLHF from a Bayesian Cognitive Modeling Perspective}




\begin{icmlauthorlist}
\icmlauthor{Khanh Nguyen}{princeton}
\end{icmlauthorlist}

\icmlaffiliation{princeton}{Department of Computer Science, Princeton University, New Jersey, USA}

\icmlcorrespondingauthor{Khanh Nguyen}{khanh.nguyen@princeton.edu}

\icmlkeywords{Theory of Mind, ToM}

\vskip 0.3in
]



\printAffiliationsAndNotice{}  

\begin{abstract}
How do language models ``think''? This paper formulates a probabilistic cognitive model called the \textit{bounded pragmatic speaker}, which can characterize the operation of different variations of language models.
Specifically, we demonstrate that large language models fine-tuned with reinforcement learning from human feedback \cite{ouyang2022training} embody a model of thought that conceptually resembles a fast-and-slow model \cite{kahneman2011thinking}, which psychologists have attributed to humans. 
We discuss the limitations of reinforcement learning from human feedback as a fast-and-slow model of thought and propose avenues for expanding this framework. 
In essence, our research highlights the value of adopting a cognitive probabilistic modeling approach to gain insights into the comprehension, evaluation, and advancement of language models.

\end{abstract}

\section{Introduction}

Large language models \cite{brown2020language,chowdhery2022palm,hoffmann2022training,zhang2022opt,scao2022bloom,touvron2023llama} have emerged as a powerful form of intelligence. 
These models demonstrate numerous traits associated with both human and superhuman intelligence.
They can engage in natural conversations with humans \citep{chatgpt}, learn from limited examples \citep{dong2022survey}, solve complex reasoning problems \citep{wei2022chain}, generate programs \citep{chen2021evaluating}, and pass exams designed for human professionals \citep{openai2023gpt4}.
Although the capabilities of large language models have been extensively documented, our understanding of the underlying cognitive mechanisms that enable these capabilities remains limited. 
By consuming a huge collection of records of human behavior and knowledge, have these models managed to think and reason like humans? Or are they merely copycats?
If neither is the case, what exactly is their ``model of thought''?  
Providing a scientific answer to these questions is crucial for dispelling unfounded speculations about large language models and guiding their future development.

\begin{figure*}[t!]
    \centering
    \includegraphics[width=\linewidth]{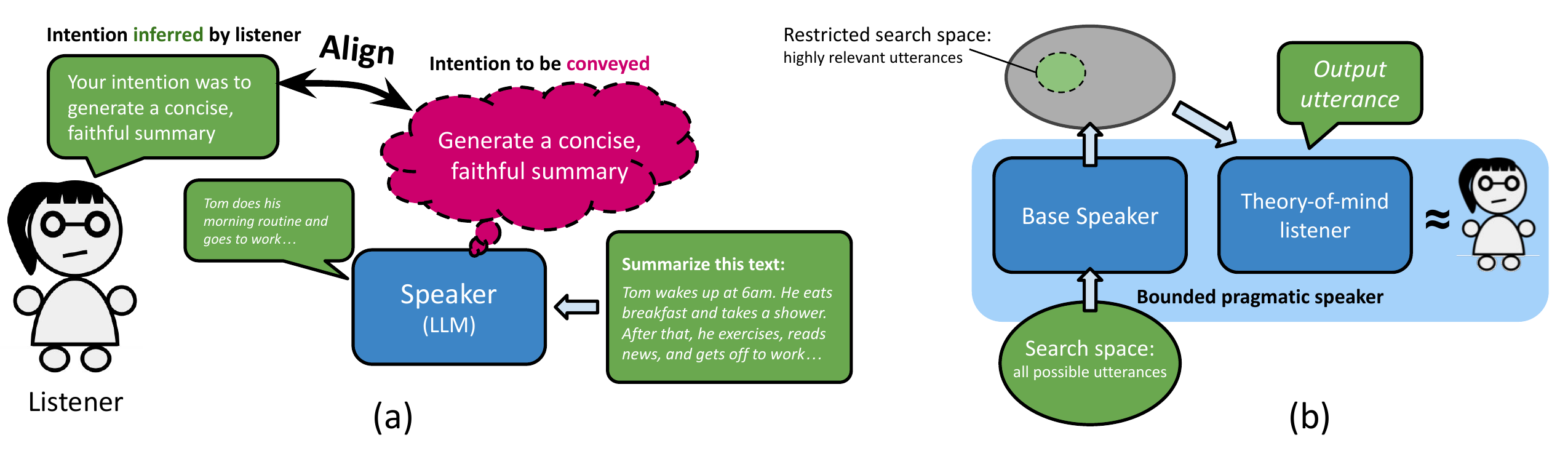}
    \caption{An overview of our proposed framework. (a) a summarization task is illustrated as a \textit{communication game}, where a speaker generates an utterance (the summary) to convey an intention (generating a good summary) given a context (the text to be summarized). The game is considered solved when the speaker presents an utterance that causes the listener to infer exactly the speaker's target intention. (b) a \textit{bounded pragmatic speaker} efficiently finds a good utterance to output by implementing a \textit{base speaker} to effectively restrict the search space, and a \textit{theory-of-mind listener} to anticipate the intention inferred by the (real) listener. }
    \label{fig:main}
\end{figure*}

In this paper, we attempt to mathematically characterize the cognitive process of large language models.
Our work is inspired by the work of \citet{mahowald2023dissociating} who propose a distinction between \textit{formal competence} (knowledge about linguistic rules and patterns) and \textit{ functional competence} (knowledge that enables pragmatic use of language) in evaluating large language models. 
To formalize this intuition, we introduce a mathematical cognitive model called the \textit{bounded pragmatic speaker} (\autoref{fig:main}), which is a generalized version of the Rational Speech Act model \cite{frank2012predicting}.
The bounded pragmatic speaker represents an agent that strives to communicate pragmatically but is constrained by its computational capacity. 
Consequently, it develops a base speaker model to effectively narrow the space of utterances to consider, and a theory-of-mind listener model to select the utterance that would trigger the desired effect in the listener's mind.
The base speaker encapsulates the formal competency of the agent, whereas the theory-of-mind listener embodies its functional competency. 
To efficiently generate utterances,
the bounded pragmatic speaker employs an approximate inference algorithm (e.g., Monte Carlo inference, variational inference, or a search algorithm).
\looseness=-1

Despite its apparent simplicity, the bounded pragmatic speaker framework provides valuable insight and guiding principles for comprehending and improving large language models. 
Its potential lies in fostering interdisciplinary connections between cognitive science, reinforcement learning, and probabilistic programming to advance the development of next-generation models. 
Our vision encompasses the creation of modular probabilistic programs that draw inspiration from human cognition and incorporate enhanced reinforcement learning techniques to achieve efficient inference.
\looseness=-1

The remainder of the paper is structured as follows. 
First, we formally define the bounded pragmatic speaker framework (\autoref{sec:bps}). 
Next, we demonstrate that a language model can be viewed as a straightforward bounded pragmatic speaker that uses its own model to serve as both a base speaker and a theory-of-mind listener (\autoref{eqn:lm_are_bps}).
This perspective on language models motivates three directions for improving them. 
In \autoref{sec:inference}, we revisit two recent extensions of large language models---pragmatic inference \cite{zhang2022coder} and reinforcement learning from human feedback \citep{ouyang2022training}---and show that they can be regarded as methods for boosting the functional competency of a bounded pragmatic speaker. 
In particular, reinforcement learning can be framed as learning a variational approximation of a bounded pragmatic speaker's distribution to allow for efficient yet pragmatic inference. 
This approach bears striking similarities with the dual model of thought proposed by \citet{kahneman2011thinking}, which is composed of a slow-thinking system that performs deep reasoning and a fast-thinking system that implements heuristics to react quickly to situations.
In the final section (\autoref{sec:future}), we argue that reinforcement learning from human feedback remains a rudimentary means of implementing a dual model of thought.
We explain the limitations of the reward function as a slow-thinking system and the inefficiency of using reinforcement learning to transfer knowledge and capabilities from the slow-thinking to the fast-thinking system.
Finally, we discuss promising ideas for devising superior alternatives.

\section{Bounded pragmatic speakers}
\label{sec:bps}

A language model can be viewed as a speaker $S(\utterance \mid \intention, \context)$ that outputs a distribution over utterances $\utterance$ to complete a task specified by a context $\context$ and an intention $\intention$.
For example, to ask a language model to generate a summary of an article, we input to the model a prompt specifying an article to be summarized (the context $\context$), and a list of the desiderata of the output summary (the intention $\intention$), and let it execute an inference algorithm to generate a summary (the utterance $\utterance$).
\looseness=-1

Generating satisfactory utterances can be formulated as solving a \textit{communication game} \citep{Lewis1969-LEWCAP-4,goodman2016pragmatic,lazaridou2016multi,wang2021calibrate}, where a speaker communicates with a listener $\reallistener(\intention \mid \utterance, \context)$ to deliver a target intention $\intention^{\star}$.
The listener can use their judgment to infer the underlying intention of an utterance. 
The objective of the speaker is to find an utterance $\utterance^{\star}$ that maximizes the probability of the listener inferring $\intention^{\star}$:
\looseness=-1
\begin{align}
    u^{\star} = \argmax_u \reallistener(\intention^{\star} \mid u, \context)
\end{align}

A communication game can be solved by an \textit{unbounded pragmatic speaker}, which has unlimited computing capacity and a perfect copy of the listener inference model. Its speaking distribution is
\begin{align}
    \unboundspeaker(u \mid \intention^{\star}, \context) \propto \reallistener(\intention^{\star} \mid u, \context)
\end{align}
With unlimited computational power, this speaker is capable of finding the optimal utterance in a reasonable amount of time by running all possible utterances through its model and selecting the one with maximum probability. 
\looseness=-1

Human and language models, however, have limited computing capacity and are better modeled as agents with bounded rationality \citep{simon1957models}.
We define a \textit{bounded pragmatic speaker} (BPS) as a speaker with bounded rationality, who possesses two capabilities: the \textit{search} capability and the \textit{pragmatic} capability.
It leverages these capabilities to efficiently approximately compute the optimal solution for the communication game.
The search capability refers to the ability to effectively narrow the search space using prior knowledge.
This capability can be formalized as having a low support probability distribution on utterances $\basespeaker(\utterance \mid \intention, \context)$, which we call the \textit{base speaker}.
The pragmatic capability allows for the construction of an approximate model of the listener $\tomlistener \approx \reallistener$, which we call the \textit{theory-of-mind listener}.
Humans are widely known to possess these two capabilities. 
We postulate the mental states of others to predict their behavior \citep{premack1978does,wimmer1983beliefs,baron1985does,gopnik1988children}.
We are also capable of quickly proposing effective candidate solutions of problems \citep{sanborn2016bayesian,vul2014one} and instantly crafting fluent and grammatically correct sentences.

Given the components $\basespeaker$ and $\tomlistener$, the speaking distribution of a BPS is defined as
\begin{align}
    \boundspeaker(\utterance \mid \intention^{\star}, \context) \propto \basespeaker(\utterance \mid \intention^{\star}, \context) \tomlistener(\intention^{\star} \mid \utterance, \context)
\label{eqn:bps}
\end{align} which is essentially a Bayesian belief update with $\basespeaker$ as the prior and $\tomlistener$ as the likelihood function.
Performing exact Bayesian inference is still intractable for this speaker. 
However, the addition of the base speaker enables it to efficiently solve communication games via approximate inference.
We will discuss this approach in \autoref{sec:inference}.

\section{Language models are bounded pragmatic speakers}
\label{eqn:lm_are_bps}

In this section, we show that any language model can be viewed as a BPS and discuss the implications arising from this viewpoint.
\looseness=-1

\subsection{Formulation}

Let $S_{\theta}(\utterance \mid \intention, \context)$ be a language model parameterized by $\theta$.
This model is equivalent to a BPS that uses $S_{\theta}$ as both its base speaker and ToM listener. 
Formally, let the base speaker $\basespeaker(\utterance \mid \intention^{\star}, c) = S_{\theta}(\utterance \mid \intention^{\star}, \context)$ and the ToM listener $\tomlistener(\intention^{\star} \mid \utterance, \context) \propto S_{\theta}(\utterance \mid \intention^{\star}, \context)$.
For any task $(\intention^{\star}, \context)$, the BPS constituted by $\basespeaker$ and $\tomlistener$, and the language model $S_{\theta}$ agree on the optimal choice:
\begin{align}
    &\argmax_u S_{\theta}(\utterance \mid \intention^{\star}, \context) \nonumber \\
    &= \argmax_u S_{\theta}(\utterance \mid \intention^{\star}, \context) S_{\theta}(\utterance \mid \intention^{\star}, \context) \nonumber \\
    &= \argmax_u \basespeaker(\utterance \mid \intention^{\star}, \context) \tomlistener(\intention^{\star} \mid \utterance, \context) 
\end{align} In other words, they exhibit identical behavior in every communication game. 
\looseness=-1

Studying this trivial BPS may not initially appear to be interesting. 
However, this perspective of a language model has conceptual value because it essentially \textbf{transforms a monolithic model into a modular one}.  
The monolithic view provides limited insight into improving language models, as their internal operations under this view are largely nebulous. 
In contrast, the BPS view establishes the connection between language models and a broader family of modular models, which offers greater interpretability as the modular structure of these models allows for independent dissection and upgrading of the modules. 
Within the BPS family of models, a (vanilla) language model can be seen as the simplest instantiation, with its modules sharing the same model. 
Therefore, it is natural to enhance language models by developing them into more sophisticated BPSs that are composed of specialized modules.
Recent developments on language models  follow this principle.

\subsection{Directions for improving a language model}
 
There are three main causes of a BPS's failure in a communication game:
\begin{enumerate}[nolistsep]
    \item \textit{Limited search capability}: the base speaker $\basespeaker$ does not assign sufficiently large probability to the optimal utterance $\utterance^{\star}$;
    \item \textit{Flawed pragmatic capability}: the ToM listener $\tomlistener$ does not accurately emulate the actual listener $\reallistener$;
    \item \textit{Inefficient or erroneous inference algorithm}:  In this case, even if both $\basespeaker$ and $\reallistener$ are perfect, the speaker is unable to find the optimal utterance within a reasonable timeframe.
\end{enumerate}

These causes point to three directions for augmenting a language model: (1) enhance its search capability (the base speaker), (2) elevate its pragmatic capability (the ToM listener), and (3) devise a more efficient and accurate inference algorithm. 
In fact, many recent advancements in language models can be categorized within these directions.
For example, training language models on vast amounts of data \citep{brown2020language} enables them to generate more relevant utterances, aligning with the objective of enhancing search capability.
Incorporating a re-ranker \cite{chiu2021innovative,cobbe2021training,zhang2022coder} or a reward function learned from human feedback \citep{stiennon2020learning,ouyang2022training} extends a model with a better ToM listener and embodies the goal of improving pragmatic capability.
Lastly, research that introduces novel decoding algorithms \cite{holtzman2019curious,li2022contrastive,lu2021neurologic} can be attributed to the direction of refining the inference algorithm.
\looseness=-1

To effectively utilize research resources, developers may want to prioritize specific directions instead of trying all of them simultaneously. 
For instance, if a language model's search capability is already sufficient, it would be more beneficial to focus on enhancing its pragmatic capability rather than the inference algorithm.  
This requires being able to diagnose the exact cause of a model's failure. 
\citet{zhao2022cognitive} propose a procedure for this purpose. 
Their idea is quite simple: to evaluate a capability of a model, comparing the model's performance on a downstream task to that of an oracle model, which is equally proficient in the evaluated capability but attains human-level proficiency in other capabilities. 
For example, to assess the pragmatic capability, one can sample a set of candidates from the model and have a human rank them, simulating an oracle model with equivalent search capability but human-level pragmatic capability. 
The performance gap between the evaluated model and the oracle model on a downstream task is then computed, with a larger gain indicating a more pronounced deficiency in the former's pragmatic capability.

\section{Improving the inference and pragmatic capability of bounded pragmatic speakers}
\label{sec:inference}

In this section, we discuss \textit{pragmatic inference} \citep{andreas2016reasoning,fried2017unified,zhang2022coder} and \textit{reinforcement learning from human feedback} (RLHF) \citep{christiano2017deep,stiennon2020learning,ouyang2022training}---two popular approaches for boosting the performance of language models. 
We will show that, under the BPS framework, these two methods essentially follow the same recipe: extending a base speaker with a ToM listener and employing a probabilistic inference algorithm to enable efficient inference.

\subsection{Pragmatic inference}

In this approach, a score function $R_{\phi} (\utterance)$ is learned and then used to evaluate a set of candidate outputs sampled from a language model $S_{\theta}$. 
The approach can be seen as performing Monte-Carlo inference on a BPS whose base speaker is the language model and ToM listener is the score function.
Concretely, let $\basespeaker(\utterance \mid \intention^{\star}, \context) = S_{\theta}(\utterance \mid \intention^{\star}, \context)$ and $\tomlistener(z^{\star} \mid \utterance, \context) \propto \exp(R_{\phi}(\utterance))$, pragmatic inference selects the output utterance $\hat u$ as follows
\begin{align}
    \hat{\utterance} 
    &= \argmax_{u \in \gU_{\textrm{cand}} \sim S_{\theta}} R_{\phi}(\utterance) \nonumber \\
    &= \argmax_{u \in \gU_{\textrm{cand}} \sim \basespeaker} \tomlistener(z^{\star} \mid \utterance, \context) \nonumber \\
    &\approx \argmax_{\utterance \in \gU} \basespeaker(\utterance \mid \intention^{\star}, \context) \tomlistener(z^{\star} \mid \utterance, \context) 
\label{eqn:prag-inf}
\end{align} where $\gU$ is the space over all possible utterances and $\gU_{\textrm{cand}}$ is a small set of candidates sampled from $\basespeaker$.
Note that the right-hand side in the last equation takes the form of a BPS (\autoref{eqn:bps}).

\subsection{Reinforcement learning from human feedback}
\label{sec:rlhf}

Variational inference is an alternative approach to approximate inference for BPS. 
It involves choosing a variational distribution $S_{\theta}$ that is efficient in inference.
The objective is to find a set of parameters $\theta$ that minimizes the KL-divergence between the variational and the approximated distributions 
\begin{align}
    &\argmin_{\theta} \text{KL}(S_{\theta} \mid\mid \boundspeaker; z^{\star}, c) 
\label{eqn:vi}
\end{align} where $\boundspeaker$ is a BPS's distribution (\autoref{eqn:bps}) and $\text{KL}(p, q; x)$ denotes the KL divergence between two conditional distributions $p(\cdot \mid x)$ and $q(\cdot \mid x)$.

RLHF is a fine-tuning approach that has been shown to effectively align large language models (LLMs) with human preference. 
The method first learns a reward function $R_{\phi}(\utterance)$ from human ratings. 
Starting with an LLM $S_0$ that was pre-trained for language modeling, the method continues to train the model to maximize the learned reward function.
A popular RLHF variant penalizes the new model for deviating too far from $S_0$, yielding the following KL-regularized objective:
\begin{align}
    \argmax_{\theta} \mathbb{E}_{\utterance \sim S_{\theta}}[R_{\phi}(\utterance)] - \beta \text{KL}(S_{\theta}; S_0) 
\label{eqn:rlhf}
\end{align}

RLHF is equivalent to applying variational inference to a BPS.
The learned reward function $R_{\phi}$ can be interpreted as a ToM listener because it predicts how the listener evaluates the alignment of an utterance with respect to an intention. 
On the other hand, the pre-trained language model represents prior knowledge and can be considered as a base speaker.
Formally, we define these components as follows: $\tomlistener(z^{\star} \mid \utterance, \context) \propto \exp(R_{\phi}(u) / \beta)$ and $\basespeaker(\utterance \mid \intention, \context) = S_0(\utterance \mid \intention, \context)$.
Then, the RLHF objective can be rewritten as
\begin{align}
    &\argmin_{\theta} -\mathbb{E}_{\utterance \sim S_{\theta}}[R_{\phi}(\utterance)] +  \beta \text{KL}(S_{\theta}; S_0) \nonumber \\
    &=\argmin_{\theta} -\mathbb{E}_{\utterance \sim S_{\theta}}[R_{\phi}(\utterance) / \beta] +  \text{KL}(S_{\theta}; S_0) \nonumber  \\
    &=\argmin_{\theta} -\mathbb{E}_{\utterance \sim S_{\theta}}[\log \tomlistener(\intention^{\star} \mid \utterance, \context) ] +  \text{KL}(S_{\theta}; S_0) \nonumber \\
    &=\argmin_{\theta} \mathbb{E}_{\utterance \sim S_{\theta}}\left[\log \frac{ S_{\theta}(\utterance \mid \intention^{\star}, \context)}{\tomlistener(\intention^{\star} \mid \utterance, \context) \basespeaker(\utterance \mid \intention^{\star}, \context)}\right] \nonumber \\
    &=\argmin_{\theta} \text{KL}(S_{\theta} \mid\mid \boundspeaker; z^{\star}, c) 
\end{align} which is exactly the variational inference's objective (\autoref{eqn:vi}).

The connection between RL and variational inference is not a new discovery (see \citet{korbak2022rl,sumers2022talk,white2020learning,levine2018reinforcement}). 
But in this context, the implication of this connection transcends the equivalence between two machine learning algorithms. 
Our finding suggests a similarity between the thinking processes of RLHF-tuned LLMs and humans, as the behaviors of both can be explained reasonably well under the BPS framework. 
This connection is surprising because it is not planned: RLHF-tuned LLMs were supposedly not inspired by computational models of human cognition. 
It can potentially bring new opportunities and perspectives to RL researchers and cognitive scientists. 
RL researchers can incorporate principles of human cognition and communication into the design of intelligent artificial agents. 
Cognitive scientists can borrow mathematical and algorithmic tools from RL to simulate more complex human behaviors.

\begin{figure*}
    \centering
    \includegraphics[width=\linewidth]{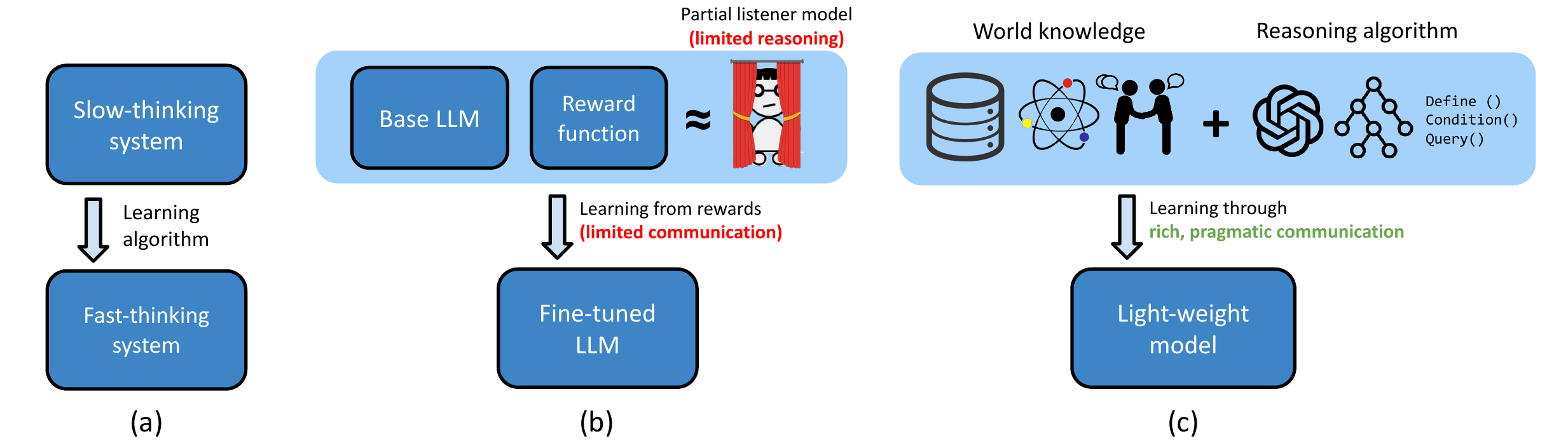}
    \caption{RLHF-tuned LLMs are instances of models that implement a \textit{dual model of thought} (a), which consists of a deliberate, methodical thinking system for rigorous reasoning (the slow-thinking system) and a quick, intuitive system for rapid decision-making (the fast-thinking system). The efficacy of the fast-thinking system can be continually enhanced by learning from the slow-thinking system. However, we argue that RLHF-tuned LLMs are still a rudimentary dual model of thought (b). The reward function fails to capture the complete reasoning capabilities of the listener, and the slow-thinking system communicates knowledge through a limited-capacity channel. We advocate for the development of a more comprehensive dual model of thought, wherein the slow-thinking system possesses extensive knowledge and profound comprehension of the physical and social world. This system would employ effective reasoning algorithms (LLMs, search algorithms, probabilistic programs, etc.) to leverage such knowledge and understanding, while facilitating efficient distillation of knowledge and capabilities into the fast-thinking system.\looseness=-1}
    \label{fig:dmot}
\end{figure*}

\section{Towards bounded pragmatic speakers with a dual model of thought}
\label{sec:future}

The variational inference approach is reminiscent of the fast-and-slow dual model of thought (DMoT) \cite{kahneman2011thinking}---a renowned theory in psychology that explains human cognition. 
A DMoT comprises of a \textit{slow-thinking system} for deep reasoning and a \textit{fast-thinking system} for fast inference. 
In the case of BPS, the speaker itself is essentially a slow-thinking system because of the expensive cost of the Bayesian inference operator.
Performing variational inference on BPS amounts to distilling the knowledge and capabilities of the slow-thinking system into a more efficient fast-thinking system using a \textit{learning algorithm}. 
In the more specific case of RLHF-tuned LLMs, the slow-thinking system is constituted by the pre-trained model (the base speaker) and the reward function (the ToM listener). 
This system is a BPS that pragmatically reasons about the real listener to make decisions. 
RL serves as the learning algorithm, constructing a fast-thinking system (the fine-tuned LLM) that agrees with the slow-thinking system in a set of situations.
If this fast thinking system robustly generalizes to new situations, it allows the LLM to communicate both efficiently \textit{and} pragmatically. 
This approach can be viewed as a form of amortized inference \cite{gershman2014amortized}, wherein inferences are ``cached'' to reduce the asymptotic cost.  
\looseness=-1

While it may not be necessary to construct an explicit fast-thinking system\footnote{For example, a Monte Carlo approach only draws a set of samples from the slow-thinking system and  considers it as an \textit{implicit} fast-thinking system.}, implementing the system as an actual machine learning model can be powerful.
High-capacity models like neural networks can potentially implement more complex algorithms than any human can design. 
Moreover, this algorithm can be continually improved by minimizing disagreement with slow-thinking system and optimizing for other intrinsic motivations (e.g., cognitive effort).
Consequently, instead of having to manually design a complex inference algorithm, one can implement a highly general model and learning algorithm, and let the optimization process automatically discover an effective inference algorithm.
\looseness=-1

DMoT is an abstract conceptualization that can manifest itself in various forms.
A slow-thinking system can be implemented in many different ways: a probabilistic model \cite{griffiths2010probabilistic}, a modular neural network \cite{corona2020modular}, a tree search algorithm \citep{anthony2017thinking,zhao2023large}, a causal graph \citep{geiger2021causal}, a program \citep{wang2023voyager}, or a language model prompted to reason and construct plans \citep{wei2022chain,ahn2022can} or engineered to represent mental states \citep{andreas-2022-language}.
A fast-thinking system can be a light-weight generative neural network.
The learning algorithm can be imitation learning, reinforcement learning, an advanced decoding algorithm \cite{lu2021neurologic}, or a learning algorithm that enables learning from rich feedback \cite{nguyen2021interactive}. 
While we could attempt all combinations, it is more useful to think about general development directions. 
In the remainder of the section, we discuss several potential directions motivated by our analysis of the fundamental limitations of RLHF as an approach to constructing a DMoT.
Our proposals are summarized in \autoref{fig:dmot}.

\subsection{Beyond reward function: slow-thinking system with strong reasoning capability}

As shown in \autoref{sec:rlhf}, an RLHF-tuned LLM defines a slow-thinking system based on a reward function $R_{\phi}(u)$, which is essentially a ToM listener
$\tomlistener(\intention^{\star} \mid \utterance, \context)$.
We argue that this function offers very limited reasoning capability.

First, the function lacks the capability of reasoning \textit{counterfactually}, because it does not model the full distribution of the true listener.
Imagine the true listener's model $\reallistener(\intention \mid \utterance, \context)$ as a matrix with rows corresponding to intentions and columns corresponding to utterances. 
The RLHF's ToM listener $\tomlistener(\intention^{\star} \mid \utterance, \context)$ captures only a single row of this matrix where $\intention = \intention^{\star}$. 
It can only predict the likelihood of an utterance $\hat \utterance$ under the target intention $z^{\star}$, but it cannot describe exactly what intentions the listener could infer from $\hat \utterance$. 
Being able to reason counterfactually is important for a model to develop a deep understanding of the consequences of its behavior, which helps it effectively adjust its behavior to achieve goals. 
For example, in a summarization task, suppose that a language model implements a ToM listener and employs it as an imaginary human judge to iteratively revise its summary before outputting a final one. 
If the model simply reasons about how a human would numerically grade its summary, it provides itself with very vague clues about how to improve the summary. 
Does a score of 6 out of 10 imply that a summary needs to be more concise or faithful, or both?
In contrast, if the model can imagine evaluation with more elaborate criteria (e.g., faithfulness, conciseness, toxicity), it can modify its summary more effectively to satisfy the real listener.
It is important to clarify that we do not claim that RLHF-tuned LLMs cannot perform counterfactual reasoning. 
In fact, they can acquire this capability by imitating records of human thoughts (see \citep{lampinen2023passive} for a general explanation).
Our argument is that the reward function does not offer counterfactual reasoning ability.
However, RLHF-tuned LLMs can still acquire this capability through other mechanisms. \looseness=-1

Second, a reward function does not capture the \textit{long-term} effect of an utterance in the world because it is only trained to predict the immediate judgment of a human on the utterance.
In reality, an utterance does not simply influence human thoughts, but those thoughts would eventually be translated into actions that alter the world. 
A safe AI agent should implement a slow-thinking system that is capable of reasoning about the long-term impact of its actions. 
When offering life advice, the agent must anticipate the potential biases that could influence users' decisions, in order to avoid recommending harmful actions. Similarly, when providing cooking recipes, it is crucial that the agent envisions the end results and considers their impact on human health, ensuring that no unintentional poison recipes are created.
These capabilities necessitate rich knowledge about the world and how humans interact in it, which is currently severely lacking in reward functions trained purely on texts and human judgements. 

These drawbacks suggest that a natural development for RLHF is to \textbf{generalize the reward function into a world model} which can postulate physical and social interactions \citep{ni2023lever,hafner2023mastering,park2023generative,yao2023tree,Wong2023FromWM} and to \textbf{develop approximate inference algorithms to reduce the cost of planning with a world model}.

\subsection{Beyond learning from rewards: transferring knowledge through rich communication}

A slow-thinking system should not only possess strong reasoning capability but also implement an algorithm for transferring its knowledge and capabilities quickly and accurately to a fast-thinking system. 
As previously shown, RLHF-tuned LLMs employ variational inference as the learning algorithm. 
This method optimizes the KL-divergence: $\text{KL}(q \mid \mid p) = \mathbb{E}_{u \sim q}[{\log \frac{q(u)}{p(u)}}]$ between a variational distribution $q$ and an approximated distribution $p$.
To augment this method, it is important to understand its basic assumptions. 
Specifically, the method assumes an efficient evaluation capability of $p$, i.e. it can swiftly and cheaply compute a score $p(u)$ for any $u$.
This minimal assumption makes variational inference applicable to a wide range of distributions but is also the root of its inefficiency.  
Because $p$ communicates with $q$ only by scores, $q$ has to propose many samples to ``guess'' the shape of $p$. 
To improve this method, we need to \textit{untie the communication bottleneck} between $q$ and $p$, giving $q$ more information than just ``how likely your sample is under my distribution''.
\looseness=-1

Suppose that we can decompose $p(u)$ into $p(u \mid z)$ and $p(z)$ such that $p(u) = \sum_z p(u \mid z)p(z)$, and that the space of $z$ is highly structured (e.g., the space of language utterances). 
We can then compute $p(z \mid u) = \sum_u p(u \mid z)p(z)$.
Now, for a sample $\hat u$ drawn from $q$, we can provide more information than just $p(\hat u)$: (i) we can offer information about $p(z)$ by sending a sample $\hat z \sim p(\cdot)$ and (ii) we can disclose information about $p(z \mid \hat u)$ by sending a sample $\hat z \sim p(\cdot \mid \hat u)$.\footnote{Note that if $z$ and $u$ has the same representation, i.e. $u = z$, then $p(u \mid z)$ or $p(z \mid u)$ is an identity mapping and $p(z)$ becomes $p(u)$. This approach is reduced to behavior cloning because $\hat z \sim p(\cdot)$ or $\hat z \sim p(\cdot \mid \hat u)$ is essentially a demonstration.} 
These pieces of information allow for the estimation of $p(z)$ and $p(u \mid z)$.
When we have good approximations of these distributions, we can fully recover $p(u)$.
This approach yields advantages if the structure of the space of $z$ enables the learning of $p(z)$ and $p(u \mid z)$ to be much more sample-efficient than directly estimating $p(u)$ via variational inference. 
For example, if $z$ is expressed in a compositional language, we can hope that if $z_1$ and $z_2$ have overlapping phrases, improving the estimation of $p(u \mid z_1)$ also refines the estimation of $p(u \mid z_2)$; similarly, we can expect that changing $p(z_1)$ to also shift $p(z_2)$.
\looseness=-1

Translated into the language of reinforcement learning, this idea basically suggests that learning can be accelerated by employing more informative and structured feedback.
Here, the term ``feedback'' refers to any piece of information about $p$ received after observing a sample $\hat u \sim q(\cdot)$.
In variational inference, feedback is a reward $p(\hat u)$.
In the approach that we suggest, feedback is $\hat z \sim p(\cdot)$ and $\hat z \sim p(\cdot \mid \hat u)$, which can be of any form.
\citet{nguyen2021interactive} demonstrate a variant of this approach where feedback is a language description. 
The authors present an algorithm with theoretical guarantees and empirically show that it is more sample-efficient than reinforcement learning baselines.
We refer the reader to the original paper for more details.

There is no free lunch: the primary challenge of this approach is to choose the type of feedback that supports fast learning and is yet inexpensive to obtain.
The superiority of natural language as a communication medium makes it a great choice for feedback conveyance.
However, collecting language feedback directly from humans is notoriously costly.
A promising future direction is to \textbf{leverage powerful LLMs to cheaply generate language feedback}.
These models possess remarkable language generation capabilities and vast common sense about the world.
With adequate fine-tuning and prompting, they can potentially be repurposed into high-quality feedback providers.
Although it is desirable to perfectly simulate human behavior, 
building models that are reliable enough to substantially reduce the amount of real human feedback would already bring immense economic values.
\looseness=-1
\section{Conclusion}

We believe that there are great opportunities for the fields of reinforcement learning, probabilistic programming, and socio-cognitive science to collaboratively contribute to the development of more capable and beneficial large language models.
In this work, we show that Bayesian models of human cognition can be used to explain the operation of large language models.
Our proposed framework represents only a simple version of the models that computational cognitive scientists have developed. 
More advanced proposals, such as hierarchical Bayesian models \cite{tenenbaum2011grow}, can potentially accommodate more complex reasoning and offer better explainability. 
It has been challenging to scale up these models to real-world problems because of their expensive inference cost. 
However, as we have shown, large language models and its learning techniques like RLHF can offer themselves as useful tools for developing more scalable Bayesian probabilistic models. 
The outcomes would yield models that not only advance the scientific pursuit of comprehending human cognition but also serve as pragmatic tools, enhancing the quality of our daily lives.
\looseness=-1

\bibliography{custom}
\bibliographystyle{tom2023}

\end{document}